\pdfoutput=1

\documentclass[letterpaper, 10 pt, conference]{ieeeconf}  % Comment this line out if you need a4paper

\usepackage[a-1b]{pdfx}
\usepackage[usenames]{xcolor}
\usepackage{graphicx}
\usepackage{subfig}
\usepackage{booktabs}
\usepackage{multirow}
\usepackage{longtable}
\usepackage{tabularx}
\usepackage[ampersand]{easylist}
\usepackage{siunitx}
\usepackage{sidecap}

\newcolumntype{Y}{>{\centering\arraybackslash}X}
\newcolumntype{R}{>{\raggedleft\arraybackslash}X}

\newcommand{\dV}{\textit{da Vinci\textsuperscript{\textregistered}}}
\newcommand{\dVSim}{\textit{da Vinci\textsuperscript{\textregistered}} Skills Simulator\texttrademark}

\title{\LARGE \bf
Real-time Teaching Cues for Automated Surgical Coaching
}

\author{Anand Malpani$^{1}$, S. Swaroop Vedula$^{2}$, Henry C. Lin$^{3}$, Gregory D. Hager$^{4}$ and Russell H. Taylor$^{5}$% <-this % stops a space
\thanks{*Anand Malpani was supported by Link Foundation Fellowship for Advanced Simulation and Training.}% <-this % stops a space
\thanks{$^{1}$Assistant Research Scientist, Malone Center for Engineering in Healthcare, Johns Hopkins University (JHU), 3400 N Charles St, Malone Hall Room 340, Baltimore, MD 21218, USA. \texttt{\small anandmalpani@jhu.edu}}%
\thanks{$^{2}$Assistant Research Professor, Malone Center for Engineering in Healthcare, JHU}%
\thanks{$^{3}$Manager, Simulation Software Research and Development, Intuitive Surgical Inc., Sunnyvale, CA, USA.}%
\thanks{$^{4}$Mandell Bellmore Professor, Department of Computer Science, JHU.}%
\thanks{$^{5}$John C. Malone Professor, Department of Computer Science, JHU.}%
}

\begin{document}

\maketitle
\thispagestyle{empty}
\pagestyle{empty}

%%%%%%%%%%%%%%%%%%%%%%%%%%%%%%%%%%%%%%%%%%%%%%%%%%%%%%%%%%%%%%%%%%%%%%%%%%%%%%%%
\begin{abstract}
With introduction of new technologies in the operating room like the \dV{} Surgical System, training surgeons to use them effectively and efficiently is crucial in the delivery of better patient care.
Coaching by an expert surgeon is effective in teaching relevant technical skills, but current methods to deliver effective coaching are limited and not scalable.
We present a virtual reality simulation-based framework for automated virtual coaching in surgical education.
We implement our framework within the \dVSim{}.
We provide three coaching modes ranging from a hands-on teacher (continuous guidance) to a hands-off guide (assistance upon request).
We present six teaching cues targeted at critical learning elements of a needle passing task, which are shown to the user based on the coaching mode. These cues are graphical overlays which guide the user, inform them about sub-par performance, and show relevant video demonstrations.
We evaluated our framework in a pilot randomized controlled trial with 16 subjects in each arm. In a post-study questionnaire, participants reported high comprehension of feedback, and perceived improvement in performance.
After three practice repetitions of the task, the control arm (independent learning) showed better motion efficiency whereas the experimental arm (received real-time coaching) had better performance of learning elements (as per the ACS Resident Skills Curriculum).
We observed statistically higher improvement in the experimental group based on one of the metrics (related to needle grasp orientation).
In conclusion, we developed an automated coach that provides real-time cues for surgical training and demonstrated its feasibility.
\end{abstract}

% ======================================================
% ======================================================
% ======================================================
\section{Introduction}
\label{ref:sec_introduction}

Since its introduction, robot-assisted minimally invasive surgery (RAMIS) has continuously revolutionized surgical procedures across disciplines -- urologic, gynecologic, head and neck, and more recently, general surgery. The \dV{} Surgical System is the predominant robotic platform to perform RAMIS procedures.
Like any other skilled activity, there is a learning curve associated with RAMIS \cite{kaul_learning_2006}.
However, there is no standardized curriculum for training surgeons in RAMIS nor a board certification along the lines of Fundamentals of Laparoscopic Surgery \cite{peters_development_2004}.
Depending on resources available at a hospital to train surgeons, there may be wide variety in technical proficiency achieved by their surgeons in RAMIS.

Outside of robot-assisted surgery, where board certifications have been established, there has been an advocacy for simulation training and competency testing by surgical educators \cite{bell_jr._why_2009}.
Even then, majority of training and learning occurs in the operating room (OR) which may not be in the best interest of the patient. Previous studies have shown that poor technical skill is associated with an increased risk of adverse patient outcomes \cite{birkmeyer_surgical_2013,fecso_effect_2017}. This motivates moving as much learning out of the OR as possible.

Advancements in technology have led to new venues for surgical training, e.g. virtual reality (VR) simulation in laboratories.
VR simulation is available round the clock and thus enables self learning among trainees.
However, adoption and usage of this technology within surgical skills training curricula has been poor \cite{van_dongen_virtual_2008}, since current VR platforms are passive and do not actively assist trainees with their learning.
We strongly believe that providing effective coaching i.e. relevant, targeted, critical and individualized, in a VR training setting can address current shortcomings of surgical training and efficiently make surgeons ``OR-ready''.

Recent studies have shown expert-based surgical coaching to be effective in skill development \cite{bonrath_comprehensive_2015,singh_randomized_2015,palter_peer_2016,soucisse_video_2016}.
In these interventions, an experienced surgeon coach engages with the trainee in an hour long session to review their performance using a video recording.
In addition to expert availability and loss of revenue to the hospital, such expert coaching is culturally limited by concerns that surgeons being coached may be perceived as incompetent or may lack autonomy \cite{mutabdzic_coaching_2015}.
We believe that automating such expert coaching can be achieved in VR simulation resolving the limitations of manual coaching.

To the best of our knowledge, there are no existing automated methods that deliver expert-like feedback for surgical training. Previous works have explored mechanisms to provide indirect feedback. Reiley et al. \cite{reiley_effects_2008} displayed a visual scale to indicate excessive force application by surgeon's instrument on objects. They showed that such feedback resulted in reduced suture breakage, lower forces, and decreased force inconsistencies among novice surgeons. Chen et al. \cite{chen_virtual_2016} demonstrated the use of virtual fixtures for development of surgical skills  in a robot-assisted suturing and knot tying simulation task. They showed that users assisted by such virtual fixtures had higher targeting accuracy and motion efficiency, and lower tearing force and number of slips. However, these mechanisms lacked feedback generation to explain mistakes and deficits in surgeon's skills. 
Teaching common errors in performance, identifying them, and providing feedback on them has proven significance and value in skill learning \cite{rogers_role_2002,gardner_embracing_2015,nathwani_relationship_2016}.
We believe that teaching and feedback are important components of effective coaching.

In this paper, we present a virtual coaching (VC) framework that delivers real-time teaching and feedback for efficient surgical skill learning in a VR setting.
We introduce the concepts of teaching cues and deficit metrics to deliver such teaching and feedback.
We present three different modes of coaching that vary in level of interventions to suit the trainee proficiency.
We describe a pilot study to test the effectiveness of our proposed VC framework in imparting surgical skills learning.
We discuss the study outcomes and potential future steps to address current limitations of the framework.

% ======================================================
% ======================================================
% ======================================================
\section{Framework}
\label{sec:framework}
Our VC framework comprises a task progress manager, three coaching modes, context-relevant teaching cues, and a module for data logging and analytics. We implemented the proposed automated surgical coach using a simulation sandbox, that is based on the open source library H3DAPI\footnote{http://h3dapi.org/}.

% ======================================================
\subsection{Task Progress Manager (TPM)}
\label{sec:task_progress_manager}
\begin{figure}
	\centering
	\includegraphics[width=0.5\textwidth]{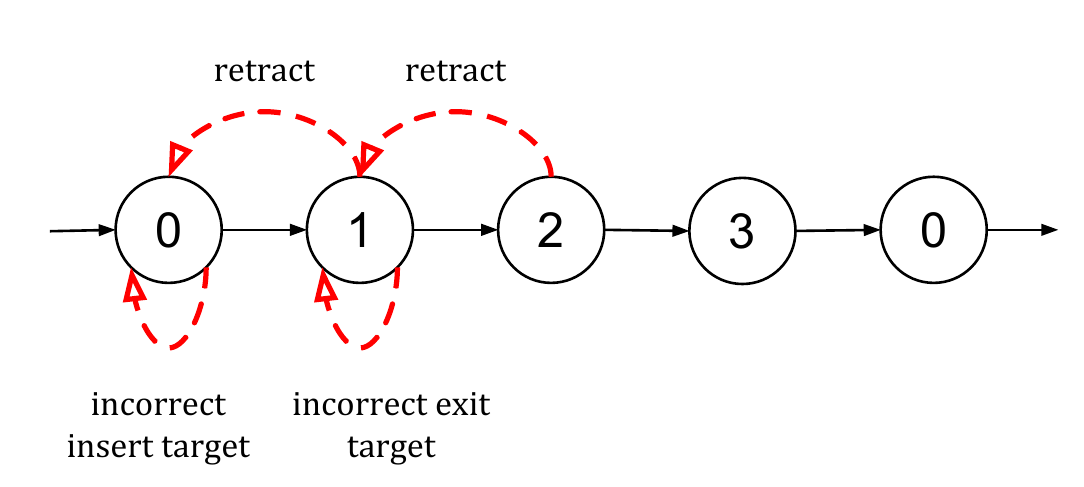}
	\caption{Task progression graph for needle passing task. The nodes represent needle passing through: none of the targets (State 0), the entry target (State 1), both of the targets (State 2), and only the exit target (State 3). The top dashed arrows indicate retraction of needle in response to errors in execution. The bottom dashed arrows indicate actions resulting in deviations from the task protocol.}
	\label{fig:needle_passing_flow}
\end{figure}

Context-relevant teaching, feedback and assessment rely on extraction of contextual information about the task flow.
We model the task flow as a directed graph.
The nodes represent task state that includes information about instruments, objects and targets. While, the edges represent a sequence of interactions that result in state transitions.
For example, in a needle passing task, current state comprises information about needle driver, needle and tissue. Interactions occur between needle driver and needle, and between needle and tissue. This sequence of interactions may result in a transition in the task flow from idle state to needle driving state.
In addition, structured surgical tasks are associated with a protocol that defines a set of rules for the user to follow for successful task execution.
For example, a needle passing task is executed through a specific order of actions such as grasping the needle, positioning it at the insertion point, driving the needle through the tissue, grasping the needle at the exit point, and rotating it out of the tissue.
TPM relies on current task state, interactions and the task-specific protocol to monitor task progress; an illustration for the needle passing task is shown in Fig. \ref{fig:needle_passing_flow}.

A VR environment provides ready and accurate access to information on task state and interactions, thereby allowing efficient monitoring of task progress. 
Our VC framework determines the correct set of parameters for the coaching modes and teaching cues, and presents them to the trainees at the right moment based on the state information obtained from the TPM.

% ======================================================
\subsection{Coaching Modes}
\label{sec:coaching_modes}
We implemented three coaching modes:
\begin{easylist}[itemize]
	& \textsc{teach} -- a complete hands-on teacher:\\
	The coach begins by demonstrating the different steps of the task, and presenting tips on them that lead to expert performance. Each of the teaching cues appear as the task progresses.
	& \textsc{metrics} -- a mentor to intervene as-needed:\\
	The coach monitors performance and intervenes only when performance falls below par based on metrics like instrument motion path length, task time, etc. Relevant teaching cues and text prompts explaining the reason for intervention appear for that particular segment only.
	& \textsc{user} -- a hands-off guide:\\
	The coach provides guidance and mentoring only when the trainee requests for it using a `help (bulb)' icon as shown in Fig. \ref{fig:teaching_cues}d. Teaching cues specific to the current task segment appear. 
\end{easylist}

% ======================================================
\subsection{Teaching Cues}
\label{sec:teaching_cues}
\begin{figure*}
	\centering
	\framebox{\parbox{6.4in}{\includegraphics[width=6.4in]{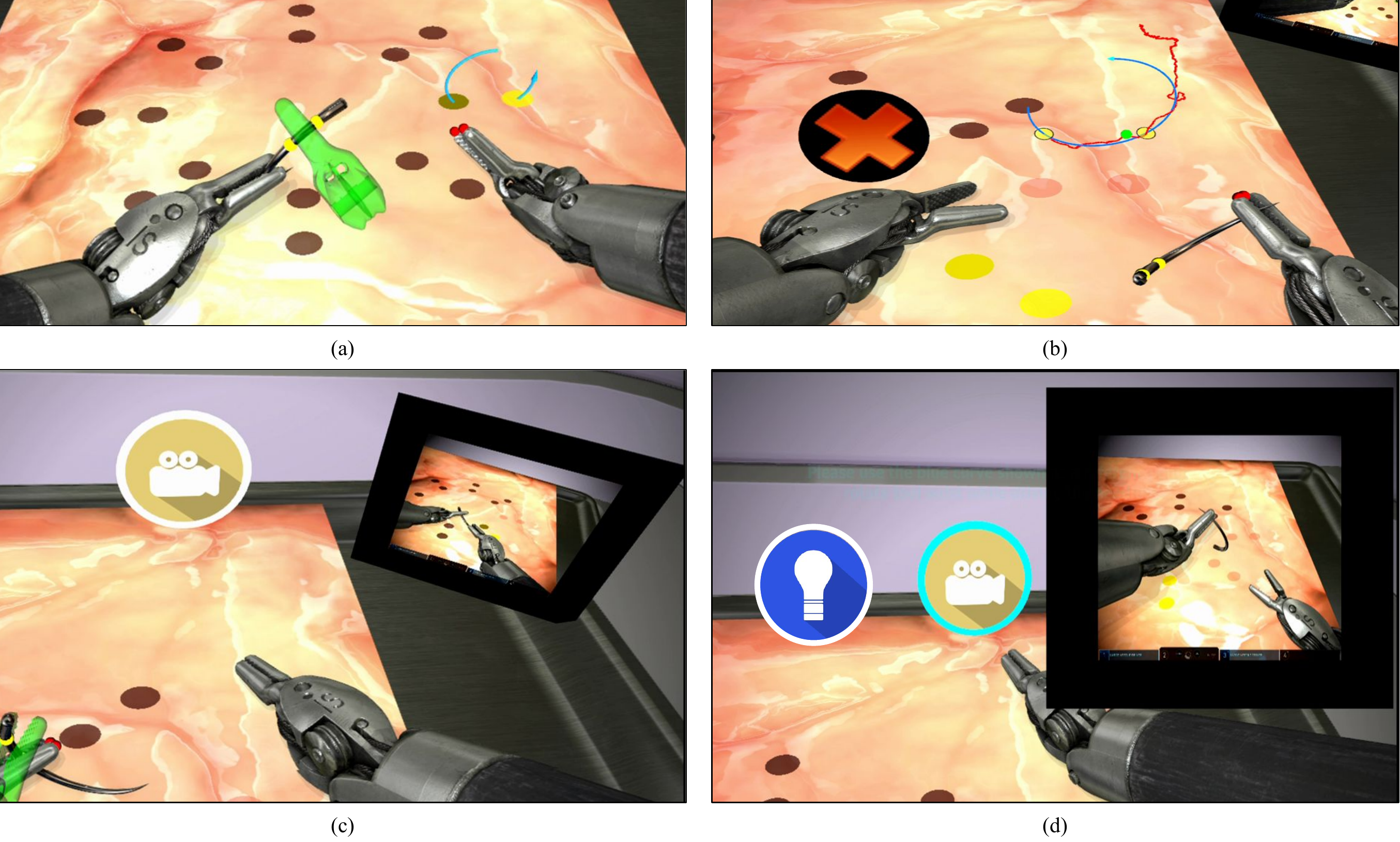}}}
	\caption{Teaching cues shown to the user in our VC framework.
	}
	\label{fig:teaching_cues}
\end{figure*}

We believe that surgical skills constitute learning elements that are critical and consequential in determining outcome of the executed skill. Our VC framework demonstrates ideal/expert behavior at such learning elements using \textbf{\textit{teaching cues}} (Fig. \ref{fig:teaching_cues}). In this work, we focus on the skill of robot-assisted needle passing (NP). The following are based on guidelines from the ACS/APDS Surgery Resident Skills Curriculum\footnote{https://www.facs.org/education/program/resident-skills} (SRSC) - modules 3, 13 and 14.

\subsubsection{Ideal Instrument Indicator}
\textbf{What}: indicates the ideal instrument (left v/s right) for performing current NP. 
\textbf{Cue}: red colored spheres at instrument tool-tip (Fig. \ref{fig:teaching_cues}a).
\textbf{How}: instrument’s initial setup pose, current entry-exit targets pose and joint limits on the robotic arms holding the instrument are used to determine the ideal instrument. In case of ambidextrous drives (either instrument is ideal), the user's handedness is used to determine the ideal instrument.
\textbf{Why}: non-ideal instrument leads to constrained ergonomics, awkward needle insertion angles and unnecessary stress on tissue.

\subsubsection{Grasp Position Guide}
\textbf{What}: indicates the suitable range of grasping regions on the needle.
\textbf{Cue}: flashing yellow spherical overlays along needle curvature (Fig. \ref{fig:teaching_cues}a).
\textbf{How}: spheres at \SI{135}{\degree} and \SI{165}{\degree} respectively, if tip of the needle is considered as \SI{0}{\degree} location. Ideally, this range may be computed using distance between entry and exit targets and needle curvature radius.
\textbf{Why}: grasping farther along the needle's body allows one smooth drive motion (bite), else leads to excessive motion as well as force exertion on the tissue.

\subsubsection{Grasp Orientation Guide}
\textbf{What}: guides the user towards an ideal grasp angle of needle with respect to instrument.
\textbf{Cue}: a light green semi-transparent copy of instrument's grippers (Fig. \ref{fig:teaching_cues}a) attached to the needle body. Guide's transparency increases as instrument approaches ideal pose, eventually disappearing.
\textbf{How}: positioned at center (\SI{150}{\degree}) of grasp position range. It is dependent on current needle drive direction and the ideal instrument suggested by the first cue.
\textbf{Why}: other grasp orientations result in excessive lateral force on tissue at insertion point, since articulation of instrument wrist gets constrained.

\subsubsection{Ideal Drive Path Overlay}
\textbf{What}: displays the ideal path for driving needle through tissue for current targets and NP.
\textbf{Cue}: cyan arc passing through current targets and pointing in the drive direction (Fig. \ref{fig:teaching_cues}a).
\textbf{How}: the ideal path has same radius as needle and height of path's center is calculated using distance between current targets and needle's radius.
\textbf{Why}: rotating the needle while driving through the tissue results in minimal lateral forces at insertion and exit targets as well as within the tissue compared to a straight motion drive.

\subsubsection{Trajectory Playback Overlay}
\textbf{What}: displays trainee’s needle tip trajectory from the previous NP attempt,
\textbf{Cue}: red curve shows needle tip trajectory along with the ideal path overlay as a cyan arc (Fig. \ref{fig:teaching_cues}b). A green sphere traces the motion at real-time speed. This is projected above the task surface for better viewing. 
\textbf{How}: trajectory of previous NP is logged along with timestamps. The orientation of cue is determined using current endoscope view direction.
\textbf{Why}: immediate visual feedback about deviation from ideal path and quality of the NP attempt may help self-reflect.

\subsubsection{Video Demonstration Overlay}
\textbf{What}: shows a video playback of an expert/ideal performance of current NP segment.
\textbf{Cue}: rectangular frame with the demonstration movie texture on an infinite playback loop (Fig. \ref{fig:teaching_cues}c, \ref{fig:teaching_cues}d). The overlay toggles between focal plane and in-situ location, if user activates the `video' icon (Fig. \ref{fig:teaching_cues}d) by bring either instrument close to it.
\textbf{How}: recorded expert (ideal) performances are used as per current NP segment being attempted.
\textbf{Why}: such demonstration enables real-time learning and shows trainees how to complete the NP attempt with success.

The initial four cues appear in the scene before the user starts driving the needle through the tissue. These cues guide the user to set up their instruments and needle in the ideal position for the current NP attempt.
Once the needle pierces the tissue, the ideal tool, grasp position and grasp orientation overlays disappear.
The ideal path overlay remains in view to guide the needle driving and it disappears once the user pulls the needle out of the exit target.
Following which, the trajectory playback overlay appears with a preset timer (10 seconds). The user is provided a `dismiss' icon (red cross) to make the playback cue disappear and continue with the task execution (Fig. \ref{fig:teaching_cues}b).
The video demonstration cue is visible throughout the task execution in the side-view location (Fig. \ref{fig:teaching_cues}c).

% ======================================================
\subsection{Error and Deficit Metrics}
\label{sec:deficit_metrics}
The teaching cues presented by our VC framework are targeted at learning elements of the needle passing task. They demonstrate ideal behavior at such elements for an overall proficient performance. 
In order to assess learning, we define \textbf{\textit{error}} and \textbf{\textit{deficit}} metrics that measure mistakes and deviations in performance at the learning elements from such ideal behavior.
We computed errors in performance using number of needle pierces in tissue, force exerted by instruments on other objects, and force exerted by needle on the tissue. An empirical force threshold was set to measure duration and count of excessive force application by instruments or needle.
The following deficit metrics were computed:
\begin{easylist}[itemize]
	& Grasp Position Deviation: average deviation in grasp location from the ideal location recommended by the \textit{Grasp Position Guide} cue (\SI{150}{\degree}),
	& Grasp Orientation Deviation: average deviation in grasp direction from the normal direction to the needle plane (indicated by \textit{Grasp Orientation Guide} cue),
	& Ideal Drive Path Deviation: average deviation of needle tip position from the ideal path indicated by the \textit{Ideal Drive Path Overlay} cue. There are two deviations: one along the depth direction (in plane) and the other in lateral direction (out of plane).
\end{easylist}
These metrics are used for evaluation of skill, as well to trigger coaching using teaching cues in the \textsc{metrics} mode.

% ======================================================
% ======================================================
% ======================================================
\section{User Study}
\label{sec:user_study}

We implemented our VC framework on the \dVSim{} (dVSim\footnote{https://www.intuitivesurgical.com/products/skills\_simulator/}), which is a portable computer that connects to the \dV{} surgeon console.

\subsection{Needle Passing Task}
We modified an existing NP task in the dVSim for this study. NP is a basic skill and core component of all surgical skill training curricula. In this study, the user is required to pass the given needle across a deformable tissue at eight locations around a circle in clockwise sequence (see Fig. \ref{fig:teaching_cues}a). The needle is to be passed from inside to the outside; the insertion targets are on the inner circle and the exit targets are on the outer circle.

\subsection{Study Design}
We conducted a randomized controlled trial (RCT) to determine the effectiveness of our VC framework on technical skill acquisition.
The study was approved by Western IRB (protocol \#20121049) and conducted at Intuitive Surgical Inc. (Sunnyvale, California).
We recruited study participants from among clinical trainers and other engineers at Intuitive Surgical, Inc. 
We randomly assigned study participants to either learning with our VC framework (experimental) or through independent, self-driven, repetitive practice (control).
All participants performed a baseline trial of the NP task.
Following this, the experimental group practiced the task under each of the three coaching modes in our VC framework -- \textsc{teach}, \textsc{metrics} and \textsc{user} (in order).
The control group independently repeated the task three times with no coaching.
Then, all participants performed a final test trial.
Following this, they responded to a post-study questionnaire on the clarity and quality of feedback, perceived effectiveness of our VC framework, and a self-assessment of their performance.
We computed performance metrics focused on time, motion efficiency, errors and deficits (Section \ref{sec:deficit_metrics}).
Motion efficiency features were based on previous works in literature \cite{dosis_synchronized_2005,jog_towards_2011}.
We compared the change in metric from baseline between the groups using a Mann-Whitney \textit{U} test for the final repetition as well as for each of the three practice repetitions.
For illustration purposes, effect size values were calculated for each metric as per the Cohen's \textit{d} statistic \cite{cohen_chapter_1977}.

% ======================================================
% ======================================================
% ======================================================
\section{Results}
\label{sec:results}
We assigned 16 participants to each arm; two participants in the experimental arm did not complete the study.
Six participants had incomplete data due to technical reasons (two in experimental and four in control).
We performed simple imputation for these incomplete data using mean for continuous measures and median for count-based ones.
Finally, we analyzed data from 30 participants (experimental: 14, control: 16).

In the post-study survey, most participants (93.3\%) perceived an improvement in their final performance relative to the baseline.
The experimental group ($\ge 85\%$) rated all but one of the teaching cues as ``intuitive'', ``clear to understand'' and ``effective for learning''.
They found the trajectory playback cue to be not intuitive, not easy to understand and not effective for learning (22\% negative rating, 22\% neutral rating).
Ninety-two percent of the experimental group felt that such feedback is essential for effective learning both in the presence and absence of a surgical educator or mentor.
The control group (68\%) felt that real-time feedback would have helped them in improving their performance. While the control group was equivocal about the effectiveness of real-time feedback (in general), 93\% of them preferred to have such feedback for themselves.

\begin{figure*}
	\centering
	\framebox{\parbox{6in}{\includegraphics[width=6in]{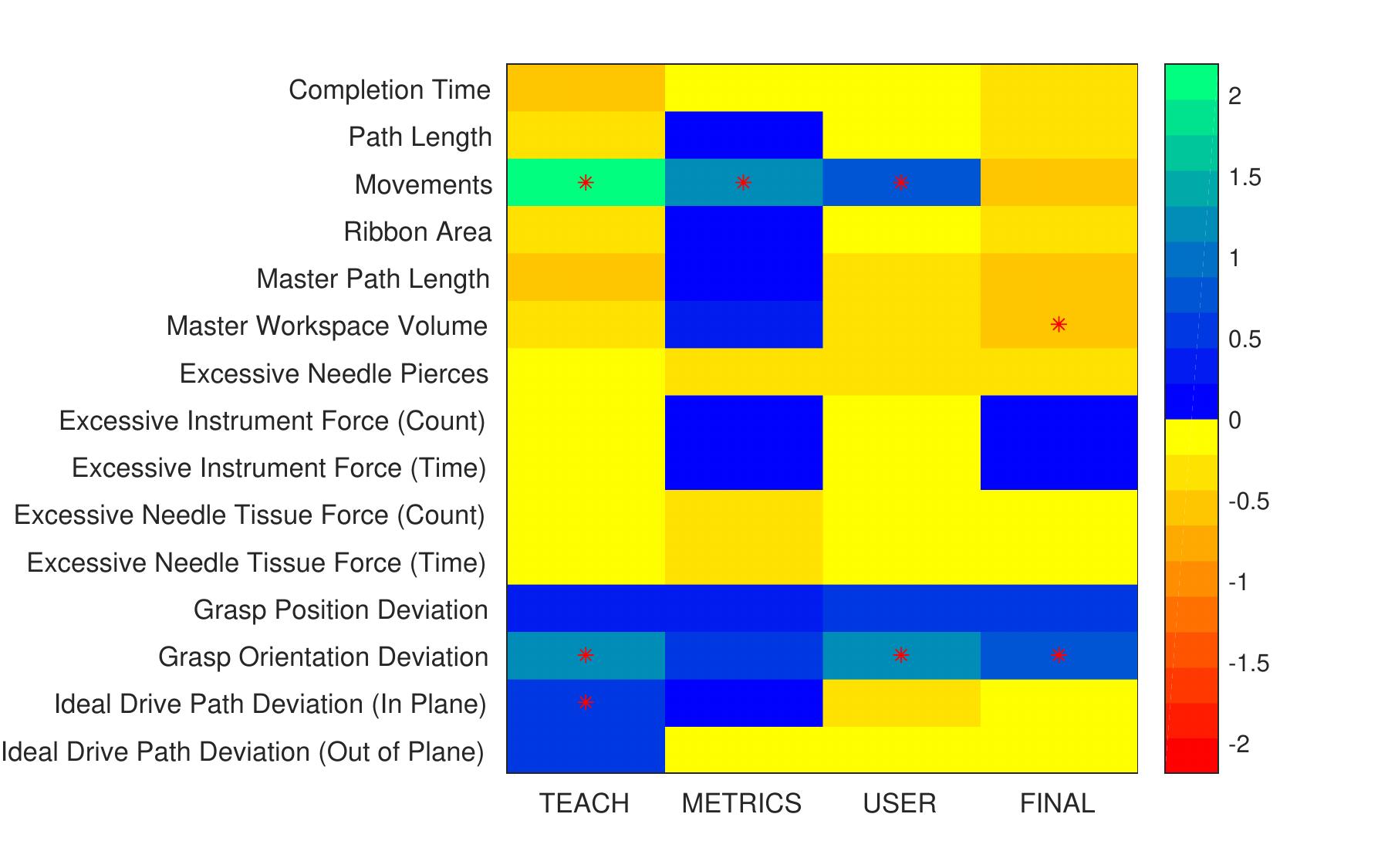}}}
	\caption[Difference between experimental and control groups: task-level performance improvement over baseline]{Comparison of performance improvement over baseline between the experimental and control groups. Each cell indicates the effect size value for the metric using Cohen's \textit{d}. Negative values (warm colors) indicate larger improvement in control group and positive values (cool colors) indicate larger improvement in experimental group. Red asterisks represent that the P-value from the Mann Whitney \textit{U} test was less than 0.05 for the particular metric and task repetition.}
	\label{fig:task_improv_btw}
\end{figure*}

We observed statistically significant difference in the improvement of performance between experimental and control groups on one metric -- Grasp Orientation Deviation (Table \ref{tab:task_improv_between}; larger negative values indicate greater learning).
Fig. \ref{fig:task_improv_btw} shows the difference (effect size values) in task-level performance improvement over the baseline between experimental and control groups.
Time and motion efficiency metrics uniformly show a higher learning in control group (warm colors), while deficit metrics show higher learning in experimental group (cool colors). Error metrics stay very close to zero indicating no difference between the two groups.
We observe that \textit{Movements} (repetitions 2, 3 and 4), \textit{Grasp Orientation Deviation} (repetitions 2,4 and 5) and \textit{Ideal Drive Path Deviation (In Plane)} (repetition 2) show statistically significantly higher performance improvement in the experimental group.
Also, the statistically higher improvement in \textit{Movements} for the exerimental group in the \textsc{teach} mode repetition becomes smaller, and eventually, the control group improvement in \textit{Movements} is higher in the \textsc{final} repetition.
Deficit metrics (lower four rows in Fig. \ref{fig:task_improv_btw}) indicate higher learning in experimental group in the \textsc{teach} mode. This learning reduces compared to the control group by the \textsc{final} repetition.

\begin{table*}
	\begin{center}
		\caption[Performance improvement from baseline on overall task execution]{Performance improvement from baseline on overall task execution at the final repetition. Experimental and control group were compared using the Mann-Whitney \textit{U} test. The numbers in the parentheses are standard deviations.}
		\label{tab:task_improv_between}
		\begin{tabularx}{0.8\textwidth}{lYYY}
			\toprule
			Metric & Experimental ($N = 14$) & Control ($N = 16$) & P value \\
			% Insert MATLAB begins
			\midrule
			Completion Time (s) & -132.71 (134.05) & -167.95 (172.90) & 0.52 \\
			Path Length (mm) & -137.04 (162.11) & -208.81 (227.47) & 0.13 \\
			Movements (count/s) & 0.25 (0.39) & 0.03 (0.45) & 0.14 \\
			Ribbon Area (mm$^2$) & -277.47 (322.11) & -427.91 (453.19) & 0.15 \\
			Master Path Length (mm) & -337.42 (376.86) & -546.20 (428.33) & 0.16 \\
			Master Workspace Volume (mm$^3$) & -294.24 (528.96) & -882.42 (1403.51) & 0.04 \\
			\midrule
			Exc. Needle Pierces & -1.50 (8.92) & -5.88 (14.57) & 0.14 \\
			Exc. Instrument Force (Count) & -1.79 (2.86) & -1.63 (5.80) & 0.53 \\
			Exc. Instrument Force (Time) (s) & -6.35 (8.07) & -5.72 (19.79) & 0.47 \\
			Exc. Needle Tissue Force (Count) & -2.93 (5.51) & -4.44 (15.59) & 0.97 \\
			Exc. Needle Tissue Force (Time) (s) & -11.81 (21.04) & -17.70 (48.45) & 0.76 \\
			\midrule
			Grasp Position Dev. (degree) & -3.73 (16.86) & 4.19 (18.32) & 0.31 \\
			\textbf{Grasp Orientation Dev. (degree)} & \textbf{-14.53 (12.99)} & \textbf{-4.22 (11.09)} & \textbf{0.04} \\
			Ideal Drive Path Dev. (In) (mm) & -0.00 (0.06) & -0.01 (0.05) & 1.00 \\
			Ideal Drive Path Dev. (Out) (mm) & -0.01 (0.05) & -0.02 (0.03) & 1.00 \\
			% Insert MATLAB ends
			\bottomrule
		\end{tabularx}
	\end{center}
	{\small Exc.: Excessive, Dev.: Deviation}
\end{table*}

% ======================================================
% ======================================================
% ======================================================
\section{Discussion and Conclusion}
\label{sec:discussion}

Our VC framework demonstrates the feasibility of an automated surgical coach that delivers relevant, targeted, critical and individualized learning.
We implemented the VC in a VR simulation sandbox in context of robot-assisted needle passing.
Our pilot study forms the basis for research and development of future automated coaching platforms in surgical skills training curricula.

Our observation of higher learning in deficit metrics and lower improvement of motion efficiency among the experimental group compared to the control group is coherent with previous findings \cite{boyle_optimising_2011,crochet_deliberate_2011,cole_randomized_2013}. For example, Singh et al. \cite{singh_randomized_2015} observed enhanced quality in performance at the expense of time and motion efficiency metrics.
Additionally, this is expected since the teaching cues are targeted at the learning elements in needle passing to improve the product quality (deficit metrics) of the task. At the same time, motion efficiency metrics are meaningful only once the task is completed with competent outcomes.

Our study was limited by the length of our VC intervention.
We exposed the experimental group to a single \textsc{teach} mode training session.
As a result, the initial improvement over learning elements (reflected by the statistically significant improvement in deficit metrics in Fig. \ref{fig:task_improv_btw}) became less significant by the \textsc{final} session.
A future study should include more number of task repetitions to effectively enable improvement in performance of trainees receiving the VC intervention.
We explored visual teaching cues in the current work. In future, haptics-based cues that use virtual fixtures \cite{chen_virtual_2016} can be added to provide a hand-over-hand guidance to demonstrate the critical points in task as well as to enable deliberate practice \cite{crochet_deliberate_2011} at segments of sub par performance. 

In summary, we addressed the current limitation of VR simulation-based training i.e. lack of expert coaching. 
We also addressed previous recommendations on development of automated tools to deliver expert surgeon-like feedback and coaching, since current approaches are limited due to scalability issues \cite{singh_randomized_2015}.
We chose to demonstrate our VC framework in VR simulation using a robot-assisted task because it affords the best opportunity to provide surgeons with context-relevant feedback with minimal overhead.
Technical skill acquired in VR simulation has been shown to subsequently transfer to bench-top simulation and the operating room \cite{larsen_efficacy_2012}.

Teaching and feedback are important components of effective surgical coaching \cite{becker_its_2009,greenberg_surgical_2015,stefanidis_developing_2016}.
Our VC framework realizes these components, using teaching cues and deficit metrics to identify and offer immediate guidance for the trainee to understand, how to correct errors and reduce them by improving their technique.
Future research should address the question of whether and to what extent, improvements in performance with automated surgical coaching transfer without attrition to the operating room, and eventually affect safety and quality of patient care.

%%%%%%%%%%%%%%%%%%%%%%%%%%%%%%%%%%%%%%%%%%%%%%%%%%%%%%%%%%%%%%%%%%%%%%%%%%%%%%%%

\section*{Acknowledgment}
We thank Anusha Balan, Sina Parastegari, Prasad and Ashwin from Intuitive Surgical Inc. (ISI, Sunnyvale, California) for their help with the dVSim. We thank Simon DiMaio (ISI) for his feedback and helpful inputs. We thank Anthony Jarc (ISI) for help with the study IRB logistics. We thank Umut and Daniel from SenseGraphics AB (Kista, Sweden) for their support related to the H3DAPI.

%%%%%%%%%%%%%%%%%%%%%%%%%%%%%%%%%%%%%%%%%%%%%%%%%%%%%%%%%%%%%%%%%%%%%%%%%%%%%%%%

%References are important to the reader; therefore, each citation must be complete and correct. If at all possible, references should be commonly available publications.

\bibliographystyle{IEEEtran}
\bibliography{IEEEabrv,2017-IROS}

\end{document}